\pdfoutput=1

\documentclass[11pt]{article}

\usepackage[final]{acl}

\usepackage{times}
\usepackage{latexsym}

\usepackage[T1]{fontenc}

\usepackage[utf8]{inputenc}

\usepackage{microtype}

\usepackage{inconsolata}

\usepackage{graphicx}
\usepackage{multirow}
\usepackage{amstext}
\usepackage{amssymb}
\usepackage{mathrsfs}
\usepackage{hyperref}       
\usepackage{url}            
\usepackage{colortbl}

\usepackage{array}
\usepackage{tabularx}
\usepackage{makecell}

\usepackage{booktabs}
\usepackage{subfigure}
\usepackage{xspace}

\usepackage{hyperref}       
\usepackage{url}            
\usepackage{colortbl}
\usepackage{subfigure}
\usepackage{array}
\usepackage{tabularx}
\usepackage{makecell}

\usepackage{algorithm}
\usepackage{algorithmicx}
\usepackage{algpseudocode} 
\usepackage{amssymb}
\usepackage{float}

\usepackage{booktabs}
\usepackage{xcolor}
\definecolor{mycolor}{rgb}{0.5,0.1,0.8} 
\usepackage{xspace}
\usepackage{xcolor}
\usepackage{tikz}
\usepackage[most]{tcolorbox}
\usepackage{svg}
\usepackage{amsmath}
\usepackage{soul}
\usepackage{wrapfig}
\usepackage{bm}
\usepackage{adjustbox}
\usepackage{enumitem}
\usepackage{pifont}
\DeclareMathOperator*{\argmin}{arg\,min}

\definecolor{hidden-red}{RGB}{205, 44, 36}
\definecolor{hidden-blue}{RGB}{194,232,247}
\definecolor{hidden-orange}{RGB}{243,202,120}
\definecolor{hidden-green}{RGB}{34,139,34}
\definecolor{hidden-pink}{RGB}{255,245,247}
\definecolor{hidden-black}{RGB}{20,68,106}
\definecolor{purple}{RGB}{144,153,196}
\definecolor{yellow}{RGB}{255,228,123}
\definecolor{tkcolor}{RGB}{224,223,255}
\definecolor{tkcolor2_back}{RGB}{249,237,238}
\definecolor{tkcolor2_frame}{RGB}{192,0,0}
\newtcolorbox{takeaways}[2][]{
	width=\columnwidth,
        toprule=0.0pt,
        leftrule=0.9pt,
        bottomrule=0.9pt,
        rightrule=0.9pt,
        arc=0pt,
	colback = tkcolor2_back, 
	colframe = tkcolor2_frame, 
	boxsep=0pt,left=7pt,right=7pt,top=4pt,bottom=4pt,
	fontupper=\linespread{0.9}\selectfont,
	title=#2,#1}

\newtcolorbox{summary}[2][]{
    width=\columnwidth,
    colback=tkcolor, 
    colframe=tkcolor,  
    boxsep=0pt,    
    arc=0pt,       
    left=10pt,     
    right=10pt,    
    top=0pt,       
    bottom=0pt,    
    fontupper=\linespread{0.9}, 
    title=#2,      
    coltitle=black,
    fonttitle=\bfseries, 
    #1              
}

\title{Token Pruning in Multimodal Large Language Models: \\ Are We Solving the Right Problem?}

\author{
    {\bf Zichen Wen}$^{1, 2, 3}$\footnotemark[1] \quad
    {\bf Yifeng Gao}$^{1, 2}$\footnotemark[1] \quad  
    {\bf Weijia Li}$^{4, 3}$ \quad
    {\bf Conghui He}$^{3}$\footnotemark[2] \quad
    {\bf Linfeng Zhang}$^{1, 2}$\footnotemark[2] \\
    \textsuperscript{1}Shanghai Jiao Tong University \quad
    \textsuperscript{2}EPIC Lab, SJTU \\
    \textsuperscript{3}Shanghai AI Laboratory \quad 
    \textsuperscript{4}Sun Yat-sen University \\
    \normalsize
    \texttt{zichen.wen@outlook.com, heconghui@pjlab.org.cn, zhanglinfeng@sjtu.edu.cn}
}

\begin{document}
\maketitle

{
\renewcommand{\thefootnote}{\fnsymbol{footnote}}
\footnotetext[1]{Equal Contribution.}
}

{
\renewcommand{\thefootnote}{\fnsymbol{footnote}}
\footnotetext[2]{Corresponding authors.}
}

\begin{abstract}
Multimodal large language models (MLLMs) have shown remarkable performance for cross-modal understanding and generation, yet still suffer from severe inference costs. 
Recently, abundant works have been proposed to solve this problem with token pruning, which identifies the redundant tokens in MLLMs and then prunes them to reduce the computation and KV storage costs, leading to significant acceleration without training. While these methods claim efficiency gains, critical questions about their fundamental design and evaluation remain unanswered: 
\emph{Why do many existing approaches underperform even compared to naive random token selection?
Are attention-based scoring sufficient for reliably identifying redundant tokens?
Is language information really helpful during token pruning?
What makes a good trade-off between token importance and duplication?
Are current evaluation protocols comprehensive and unbiased?} 
The ignorance of previous research on these problems hinders the long-term development of token pruning. 
In this paper, we answer these questions one by one, providing insights into the design of future token pruning methods. 

\end{abstract}

\section{Introduction}
Multi-modal language models (MLLMs) \citep{huang2023languageneedaligningperception, driess2023palmeembodiedmultimodallanguage, liu2024visual, Qwen-VL}, especially the vision-language models have demonstrated powerful effectiveness in various tasks. However, the extremely high computational and storage costs have limited the application of MLLMs in real-time applications, which is caused by not only the enormous parameters inherited from LLMs but also a large number of tokens from the large visual information such as high-resolution images and multi-frame videos.

To solve this problem, abundant efforts have been made in \textbf{token pruning} \citep{chen2024image, zhang2024sparsevlm, liu2024multi,liu2025shifting,liu2025vidcom2}, which aims to reduce the number of input tokens in MLLMs. Usually, token pruning methods first introduce a carefully-designed criterion to measure the importance of a vision token, and then prune the redundant tokens, or merge the redundant tokens into fewer tokens. As a result, the following computation of the pruned tokens or the merged tokens can be removed or reduced, bringing efficiency in both computation and storage. For instance, some recent works show that more than 70\% tokens can be pruned with a tolerant loss in accuracy \citep{chen2024image}. Most attractively, thanks to the natural ability of MLLMs to process tokens in different lengths, token pruning can be applied to most existing MLLMs with no need for additional training, and thus attracts great attention from both academic researchers and industrial developers. 

\begin{figure}[!t]
    \centering
    \includegraphics[width=1.0\linewidth]{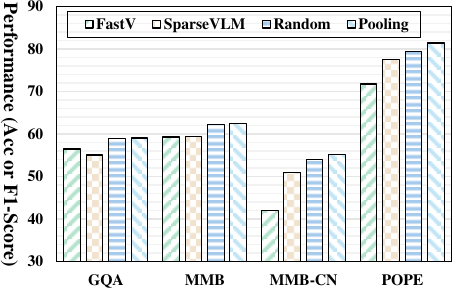}
    \caption{\textbf{Comparison between FastV, SparseVLM, and naive baselines.} On several common datasets, the performance of FastV and SparseVLM is even worse than random token dropping and pooling.}
    \vspace{-0.7cm}
    \label{fig:intro_hist}
\end{figure}

However, despite the popularity of token pruning, numerous foundational questions have long been overlooked and remain largely unexplored, giving rise to several surprising phenomena. For instance, Figure \ref{fig:intro_hist} demonstrates the comparison between two classical token pruning methods including FastV \citep{chen2024image} and SparseVLM \citep{zhang2024sparsevlm}, and two naive baselines, including random token selection and direct average pooling on tokens. \textbf{Surprisingly, the two baselines outperform the two well-designed token pruning methods in most benchmarks by a clear margin}. This counterintuitive phenomenon may demonstrate that the current understanding of so-called important tokens is far away from the truth. Unfortunately, most recent works just focus on purchasing higher performance, while ignoring these questions, which may hinder the long-term development of token pruning.

In this paper, we have conducted massive experiments and analyses to dive into the fundamental problems of token pruning, with the main takeaways as follows.
\begin{itemize}[leftmargin=10pt, topsep=0pt, itemsep=1pt, partopsep=1pt, parsep=1pt]
    \item Attention-based token selection methods suffer from position bias, where vision tokens in the later positions are more likely to be retained. Reducing position bias in these methods can benefit their performance.
    \item Language information is helpful in token pruning only when a given task strongly correlates with the language information.
    \item Both the importance and uniqueness (low similarity) of tokens
    have a significant influence on the performance of token pruning and their influence varies from different tasks.
    \item FLOPs and the number of retained tokens are unreliable metrics for token pruning methods. Compatibility with hardware has a significant influence on real acceleration performance.
    \item Training-aware token pruning which directly merges tokens in spatially adjacent positions may bring more benefits than carefully-designed training-aware pruning.
\end{itemize}

We hope that this paper can provide insights into the future design of token pruning, and correct the long-neglected evaluation issues in this field.

\section{Related Work}
\subsection{Multimodal Large Language Models}

The remarkable success of large language models (LLMs) \citep{radford2019language, brown2020language,wen2024aidbench} has spurred a growing trend of extending their advanced reasoning capabilities to multi-modal tasks, leading to the development of vision-language models (VLMs) \citep{huang2023languageneedaligningperception, driess2023palmeembodiedmultimodallanguage, liu2024visual, Qwen-VL}. These VLMs typically consist of a visual encoder \citep{radford2021learning} that serializes input image representations and an LLM responsible for text generation. To enable the LLM to process visual inputs, an alignment module is employed to bridge the gap between visual and textual modalities. This module can take various forms, such as an MLP projector or a more complex query-based network. While this integration allows the LLM to gain visual perception, it also introduces significant computational challenges due to the long sequences of visual tokens.

Moreover, existing VLMs often exhibit limitations, such as visual shortcomings or hallucinations, which hinder their performance. Efforts to enhance VLM capabilities by increasing input image resolution have further exacerbated computational demands. For instance, encoding higher-resolution images results in a substantial increase in the number of visual tokens. A model like LLaVA-1.5 \citep{liu2024improved} generates 576 visual tokens for a single image, while its successor, LLaVA-NeXT \citep{liu2024llavanext}, produces up to 2880 tokens at double the resolution, far exceeding the length of typical textual prompts.
Optimizing the inference efficiency of VLMs is thus a critical task to facilitate their deployment in real-world scenarios with limited computational resources.

\subsection{Visual Token Compression}

Visual tokens are often significantly more numerous than text tokens, with higher spatial redundancy and lower information density. To address this issue, various methods have been proposed for reducing visual token counts in vision language models. For instance, some approaches modify model components, such as using context tokens in Q-Former \citep{li2023llama} or applying adaptive pooling at the patch level, but these typically require additional training and increase computational costs. Other techniques, like Token Merging (ToMe) \citep{bolya2022tome} and FastV \citep{chen2024image}, focus on reducing tokens without retraining by merging tokens or selecting important ones based on attention scores. SparseVLM \cite{zhang2024sparsevlm} incorporates text guidance through cross-modal attention to refine token selection. However, these methods often overlook hardware acceleration compatibility and fail to account for token duplication alongside token importance. Furthermore, while token pruning has been extensively explored in natural language processing and computer vision to improve inference efficiency, its application to VLMs remains under-explored. Existing pruning strategies, such as those in FastV and SparseVLM, rely on text-visual attention within large language models (LLMs) to evaluate token importance, which may not align well with actual visual token relevance.

\section{Benchmarking}



We begin by presenting the datasets, models, and pruning methods included in our study, along with the rationale behind the selection.
Next, we outline the experimental setup and provide guidance on interpreting the results reported in our study. Finally, we analyze the findings, emphasizing notable patterns and offering insights that may inform future research in this area.

\subsection{Models}
We selected several representative MLLMs, including LLaVA-1.5-7B and 13B \citep{liu2024improved}, LLaVA-Next-7B \citep{liu2024llavanext}, and Qwen2-VL \citep{wang2024qwen2} series (7B-Instruct and 72B-Instruct).
LLaVA-1.5-7B integrates CLIP and LLaMA for vision-language alignment via end-to-end training,
employing MLP connectors to fuse visual-text features for multimodal reasoning.
LLaVA-Next-7B enhances data efficiency and inference robustness with dynamic resolution and hierarchical feature integration, improving fine-grained visual understanding.
Qwen2-VL series excel in high-resolution input processing and instruction-following,
supporting complex tasks like document analysis and cross-modal in-context learning through unified vision-language representations.

\subsection{Datasets}

To evaluate the impact of pruning on different tasks, we selected a diverse set of datasets, including visual understanding tasks including GQA \citep{hudson2019gqa}, MMBench (MMB) \citep{liu2025mmbench}, MME \citep{fu2023mme}, POPE \citep{li2023evaluating}, ScienceQA \citep{lu2022learn}, VQA$^{\text{V2}}$ (VQA V2) \citep{goyal2017making} and VQA$^{\text{Text}}$ (TextVQA) \citep{singh2019towards}, grounding task RefCOCO \citep{yu2016modelingcontextreferringexpressions, mao2016generationcomprehensionunambiguousobject} and object retrieval task Visual Haystack \citep{wu2025visual}, . We briefly introduce these datasets in Table \ref{tab:datasets}.



\subsection{Token Pruning Method}

To rigorously evaluate the properties of visual token pruning, we select three representative and high-performing methods: FastV \citep{chen2024image}, SparseVLM \cite{zhang2024sparsevlm}, and MustDrop \citep{liu2024multi}.
FastV \cite{chen2024image} optimizes computational efficiency by learning adaptive attention patterns in early layers and pruning low-attention visual tokens post-layer 2 of LLMs, effectively reducing redundancy.
SparseVLM \citep{zhang2024sparsevlm} introduces a text-guided, training-free pruning mechanism that leverages self-attention matrices between text and visual tokens to assess importance. It maximizes sparsity while preserving semantically relevant tokens without additional parameters or fine-tuning.
MustDrop \citep{liu2024multi} addresses token redundancy across the entire model lifecycle. It merges spatially similar tokens during vision encoding, employs text-guided dual-attention filtering in prefilling, and implements output-aware KV cache compression during decoding. This multi-stage approach ensures balanced retention of critical tokens while enhancing inference efficiency.
These methods exemplify diverse strategies for token pruning, spanning adaptive attention, text-guided sparsity, and lifecycle-aware optimization.

\renewcommand{\multirowsetup}{\centering}
\definecolor{mygray}{gray}{.92}
\definecolor{ForestGreen}{RGB}{34,139,34}
\newcommand{\fg}[1]{\mathbf{\mathcolor{ForestGreen}{#1}}}
\definecolor{Forestred}{RGB}{220,50,50}
\newcommand{\fr}[1]{\mathbf{\mathcolor{Forestred}{#1}}}
\begin{table*}[t]
    \centering
    \vspace{-2mm}
    \setlength{\tabcolsep}{2.3pt}
    \renewcommand{\arraystretch}{1.1}
    \footnotesize
	\centering
    \begin{tabular}{c  c| c c c c c c c c | >{\centering\arraybackslash}p{2cm}}
        \toprule[1.5pt]
        & \textbf{Method} & \textbf{GQA} & \textbf{MMB} & \textbf{MMB-CN} & \textbf{MME} & \textbf{POPE} & \textbf{SQA} & \textbf{VQA}$^{\text{Text}}$ & \textbf{VizWiz} & \makecell[c]{\textbf{Avg}.}\\
        \hline
        \rowcolor{mygray}
        \multicolumn{11}{c}{\textit{Upper Bound, 576 Tokens} \ $\textbf{(100\%)}$}\\
        & \multirow{1}*{\textcolor{gray}{Vanilla}} & \textcolor{gray}{61.9} & \textcolor{gray}{64.7} & \textcolor{gray}{58.1} & \textcolor{gray}{1862} & \textcolor{gray}{85.9} & \textcolor{gray}{69.5} & \textcolor{gray}{58.2} & \textcolor{gray}{50.0} & \multirow{1}*{\textcolor{gray}{100\%}} \\
        \hline

        \rowcolor{mygray}
        \multicolumn{11}{c}{\textit{Retain 144 Tokens} \ $\fg{(\downarrow 75.0\%)}$}   \\

        & Random & 59.0 & 62.2 & 54.1 & 1736 & 79.4 & 67.8 & 51.7 & \textbf{51.9} & 95.0\% {\color[HTML]{18A6C2} \scriptsize (-5.0\%)} \\
        & Pooling & 59.1 & \textbf{62.5} & \textbf{55.2} & \textbf{1763} & \textbf{81.4} & 69.1 & 53.4 & \textbf{51.9} & 96.4\% {\color[HTML]{18A6C2} \scriptsize (-3.6\%)}  \\
        & Window FastV & \textbf{59.2} & 59.3 & 51.0 & 1737 & 80.3 & 66.4 & 50.8 & 50.3 & 93.2\% {\color[HTML]{18A6C2} \scriptsize (-6.8\%)} \\
        \hline
        & Vanilla FastV & 56.5 & 59.3 & 42.1 & 1689 & 71.8 & 65.3 & 53.6 & 51.3 & 89.8\% {\color[HTML]{18A6C2} \scriptsize (-10.2\%)}  \\
        & Reverse FastV & 49.9 & 36.9 & 26.4 & 1239 & 59.8 & 60.9 & 36.9 & 48.4 & 70.8\% {\color[HTML]{18A6C2} \scriptsize (-29.2\%)} \\
        & SparseVLM & 55.1 & 59.5 & 51.0 & 1711 & 77.6 & \textbf{69.3} & \textbf{54.9} & 51.4 & 93.5\% {\color[HTML]{18A6C2} \scriptsize (-6.5\%)}   \\

        \rowcolor{mygray}
        \multicolumn{11}{c}{\textit{Retain 64 Tokens} \ $\fg{(\downarrow 88.9\%)}$}  \\
        & Random & \textbf{55.9} & \textbf{58.1} & \textbf{48.1} & 1599 & 70.4 & 66.8 & 48.2 & \textbf{51.6} & 89.1\% {\color[HTML]{18A6C2} \scriptsize (-10.9\%)}    \\
        & Pooling & 54.2 & 56.0 & 46.0 & 1545 & 71.2 & \textbf{67.2} & 49.4 & 49.9 & 87.6\% {\color[HTML]{18A6C2} \scriptsize (-12.4\%)}  \\
        & Window FastV & 55.8 & 56.2 & 41.2 & \textbf{1630} & 72.6 & 66.3 & 47.8 & 50.0 & 87.2\% {\color[HTML]{18A6C2} \scriptsize (-12.8\%)} \\
        \hline
        & Vanilla FastV & 46.1 & 47.2 & 38.1 & 1255 & 58.6 & 64.9 & 47.8 & 50.8 & 78.2\% {\color[HTML]{18A6C2} \scriptsize (-21.8\%)}  \\
        & Reverse FastV & 44.6 & 24.0 & 15.7 & 1114 & 45.2 & 60.8 & 35.9 & 48.4 & 61.8\% {\color[HTML]{18A6C2} \scriptsize (-38.2\%)} \\
        & SparseVLM & 52.7 & 56.2 & 46.1 & 1505 & \textbf{75.1} & 62.2 & \textbf{51.8} & 50.1 & 87.3\% {\color[HTML]{18A6C2} \scriptsize (-12.7\%)}  \\

        \bottomrule[1.5pt]
	\end{tabular}
    \caption{\textbf{Performance Comparison of LLaVA-1.5-7B with Different Token Retention Strategies.} Reverse FastV is a variant of the FastV that retains tokens with the smallest attention scores.}
    \label{tab:random_and_pooling}
\end{table*}

\begin{table*}[t]
    \centering
    \setlength{\tabcolsep}{2.3pt}
    \renewcommand{\arraystretch}{1.1}
    \footnotesize
	\centering
    \begin{tabular}{c  c| c c c c c c c c | >{\centering\arraybackslash}p{2cm}}
        \toprule[1.5pt]
        & \textbf{Method} & \textbf{GQA} & \textbf{MMB} & \textbf{MMB-CN} & \textbf{MME} & \textbf{POPE} & \textbf{SQA} & \textbf{VQA}$^{\text{Text}}$ & \textbf{VizWiz} & \makecell[c]{\textbf{Avg}.}  \\
        \hline
        \rowcolor{mygray}
        \multicolumn{11}{c}{\textit{Upper Bound, 576 Tokens} \ $\textbf{(100\%)}$}\\
        & \multirow{1}*{\textcolor{gray}{Vanilla}} & \textcolor{gray}{63.3} & \textcolor{gray}{68.9} & \textcolor{gray}{62.3} & \textcolor{gray}{1818} & \textcolor{gray}{85.9} & \textcolor{gray}{72.8} & \textcolor{gray}{61.3} & \textcolor{gray}{56.6} & \multirow{1}*{\textcolor{gray}{100\%}} \\
        \hline

        \rowcolor{mygray}
        \multicolumn{11}{c}{\textit{Retain 144 Tokens} \ $\fg{(\downarrow 75.0\%)}$}   \\

        & Random & 60.3 & \textbf{65.9} & \textbf{58.3} & \textbf{1767} & 80.6 & \textbf{71.4} & 54.3 & 57.6 & 95.5\% {\color[HTML]{18A6C2} \scriptsize (-4.5\%)} \\
        & Pooling & \textbf{60.6} & 65.5 & 57.7 & 1742 & \textbf{83.6} & 71.3 & \textbf{56.3} & 56.6 & 95.8\% {\color[HTML]{18A6C2} \scriptsize (-4.2\%)}  \\
        & Window FastV & 59.5 & 65.5 & 57.7 & 1674 & 82.8 & 57.2 & 48.0 & \textbf{60.9} & 91.8\% {\color[HTML]{18A6C2} \scriptsize (-8.2\%)} \\
        \hline
        & Vanilla FastV & 57.7 & 53.9 & 46.8 & 1633 & 79.3 & 57.0 & 51.0 & 60.3 & 86.9\% {\color[HTML]{18A6C2} \scriptsize (-13.1\%)}  \\
        & SparseVLM & 57.9 & 63.8 & 55.8 & 1704 & 81.1 & 69.9 & 43.9 & 56.3 & 91.1\% {\color[HTML]{18A6C2} \scriptsize (-8.9\%)}   \\

        \rowcolor{mygray}
        \multicolumn{11}{c}{\textit{Retain 64 Tokens} \ $\fg{(\downarrow 88.9\%)}$}  \\
        & Random & \textbf{57.5} & \textbf{62.6} & 54.4 & \textbf{1681} & 73.8 & 70.9 & 50.6 & 57.5 & 91.1\% {\color[HTML]{18A6C2} \scriptsize (-8.9\%)}    \\
        & Pooling & 55.4 & 58.3 & 51.4 & 1552 & 74.0 & \textbf{72.0} & \textbf{52.3} & 51.4 & 87.7\% {\color[HTML]{18A6C2} \scriptsize (-12.3\%)}  \\
        & Window FastV & 56.7 & 58.3 & 51.4 & 1599 & \textbf{76.5} & 55.4 & 43.2 & \textbf{59.5} & 85.7\% {\color[HTML]{18A6C2} \scriptsize (-14.3\%)} \\
        \hline
        & Vanilla FastV & 53.7 & 50.9 & 42.1 & 1567 & 69.3 & 56.8 & 47.1 & 59.2 & 81.6\% {\color[HTML]{18A6C2} \scriptsize (-18.4\%)}  \\
        & SparseVLM & 50.6 & 61.3 & \textbf{54.8} & 1402 & 65.0 & 69.0 & 22.7 & 54.5 & 79.7\% {\color[HTML]{18A6C2} \scriptsize (-20.3\%)}  \\

        \bottomrule[1.5pt]
	\end{tabular}
    \caption{\textbf{Performance Comparison of LLaVA-1.5-13B with Different Token Retention Strategies.}}
    \label{tab:random_and_pooling_13B}
     \vspace{-2mm}
\end{table*}

\begin{table}[t]
    \centering
    \setlength{\tabcolsep}{2.0pt}
    \renewcommand{\arraystretch}{1.15}
    \footnotesize
    \centering
    \scalebox{0.85}{
    \begin{tabular}{c  c| c c c c c | c}
        \toprule[1.5pt]
        & \textbf{Method} & \textbf{GQA} & \textbf{MME} & \textbf{POPE} & \textbf{SQA} & \textbf{VQA}$^{\text{Text}}$ & \makecell[c]{\textbf{Avg}.}  \\
        \hline
        \rowcolor{mygray}
        \multicolumn{8}{c}{\textit{Upper Bound, 576 Tokens} \ $\textbf{(100\%)}$}\\
        & \multirow{1}*{\textcolor{gray}{Vanilla}} & 65.3 & 2521 & 87.4 & 91.6 & 82.8 & 100\% \\

        \rowcolor{mygray}
        \multicolumn{8}{c}{\textit{Retain 144 Tokens} \ $\fg{(\downarrow 75.0\%)}$}   \\

        & Random & 63.9 & \textbf{2476} & 87.1 & 85.7 & 74.0 & 95.7\% {\color[HTML]{18A6C2} \scriptsize (-4.3\%)} \\
        & Pooling & 63.1 & 2463 & 86.9 & \textbf{86.9} & 75.1 & 95.9\% {\color[HTML]{18A6C2} \scriptsize (-4.1\%)} \\
        & Window FastV & \textbf{64.2} & 2445 & \textbf{88.5} & 85.9 & 75.4 & 96.3\% {\color[HTML]{18A6C2} \scriptsize (-3.7\%)} \\
        & Vanilla FastV & 56.5 & 2219 & 80.9 & 85.3 & \textbf{75.6} & 90.3\% {\color[HTML]{18A6C2} \scriptsize (-9.7\%)}  \\

        \rowcolor{mygray}
        \multicolumn{8}{c}{\textit{Retain 64 Tokens} \ $\fg{(\downarrow 88.9\%)}$}  \\
        & Random & \textbf{61.9} & \textbf{2394} & 85.5 & 79.3 & 64.5 & 90.4\% {\color[HTML]{18A6C2} \scriptsize (-9.6\%)}   \\
        & Pooling & \textbf{61.9} & 2391 & 84.3 & 81.1 & 65.9 & 90.9\% {\color[HTML]{18A6C2} \scriptsize (-9.1\%)} \\
        & Window FastV & \textbf{61.9} & 2377 & \textbf{85.6} & \textbf{83.8} & 65.8 & 91.6\% {\color[HTML]{18A6C2} \scriptsize (-8.4\%)} \\
        & Vanilla FastV & 55.7 & 2089 & 78.7 & 83.3 & \textbf{66.8} & 86.0\% {\color[HTML]{18A6C2} \scriptsize (-14.0\%)}  \\

        \bottomrule[1.5pt]
	\end{tabular}}
     \vspace{-2mm}
    \caption{\textbf{Performance Comparison of Qwen2-VL-72B with Different Token Retention Strategies.}}
    \label{tab:random_and_pooling_qwen2vl}
     \vspace{-4mm}
\end{table}

    
\begin{figure}[!ht]
    \vspace{-3mm}
    \centering
    \subfigure[Token Frequency]{
    \includegraphics[width=0.473\linewidth]{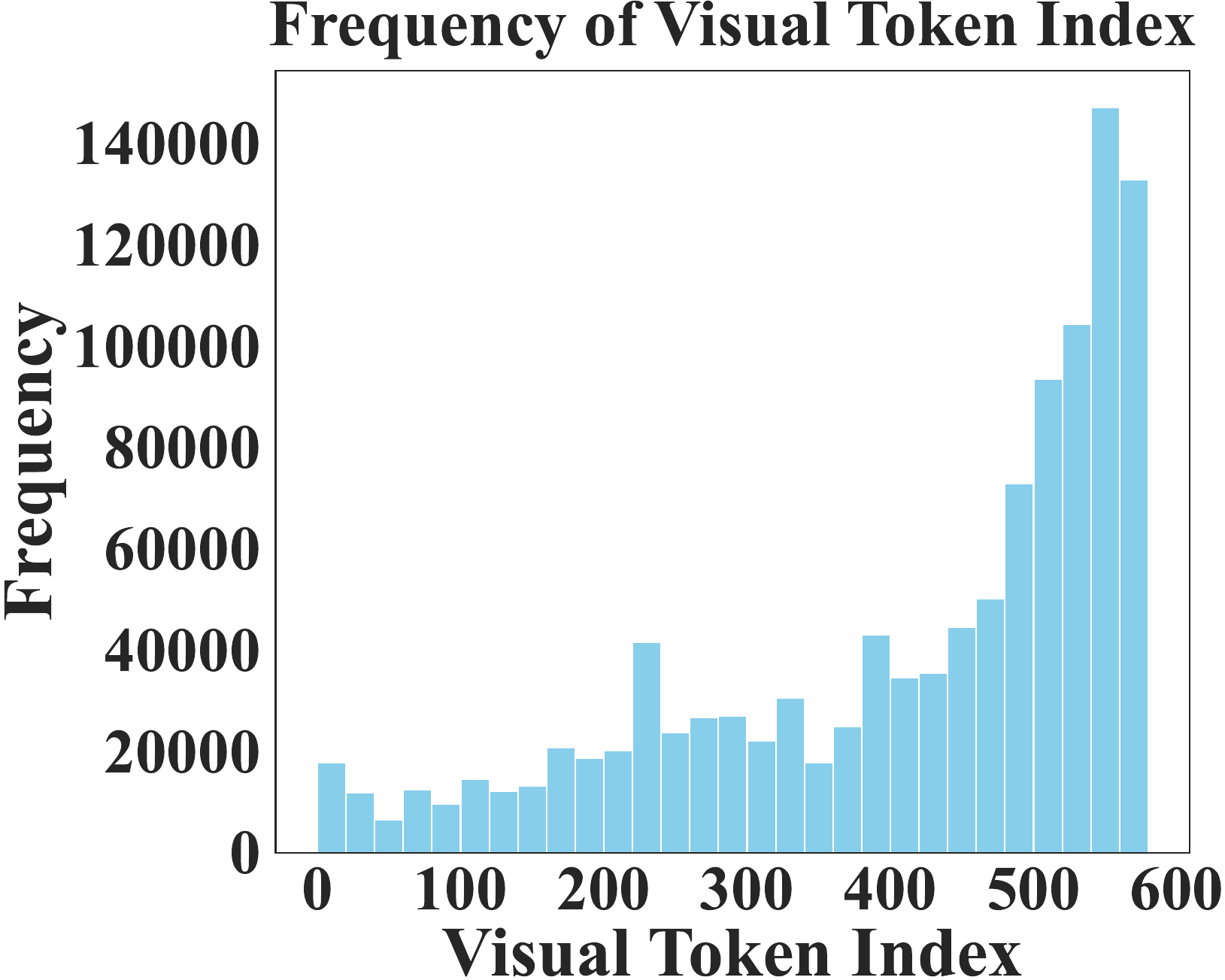}
    }
    \subfigure[Token Attention]{
        \includegraphics[width=0.465\linewidth, height=0.38\linewidth]{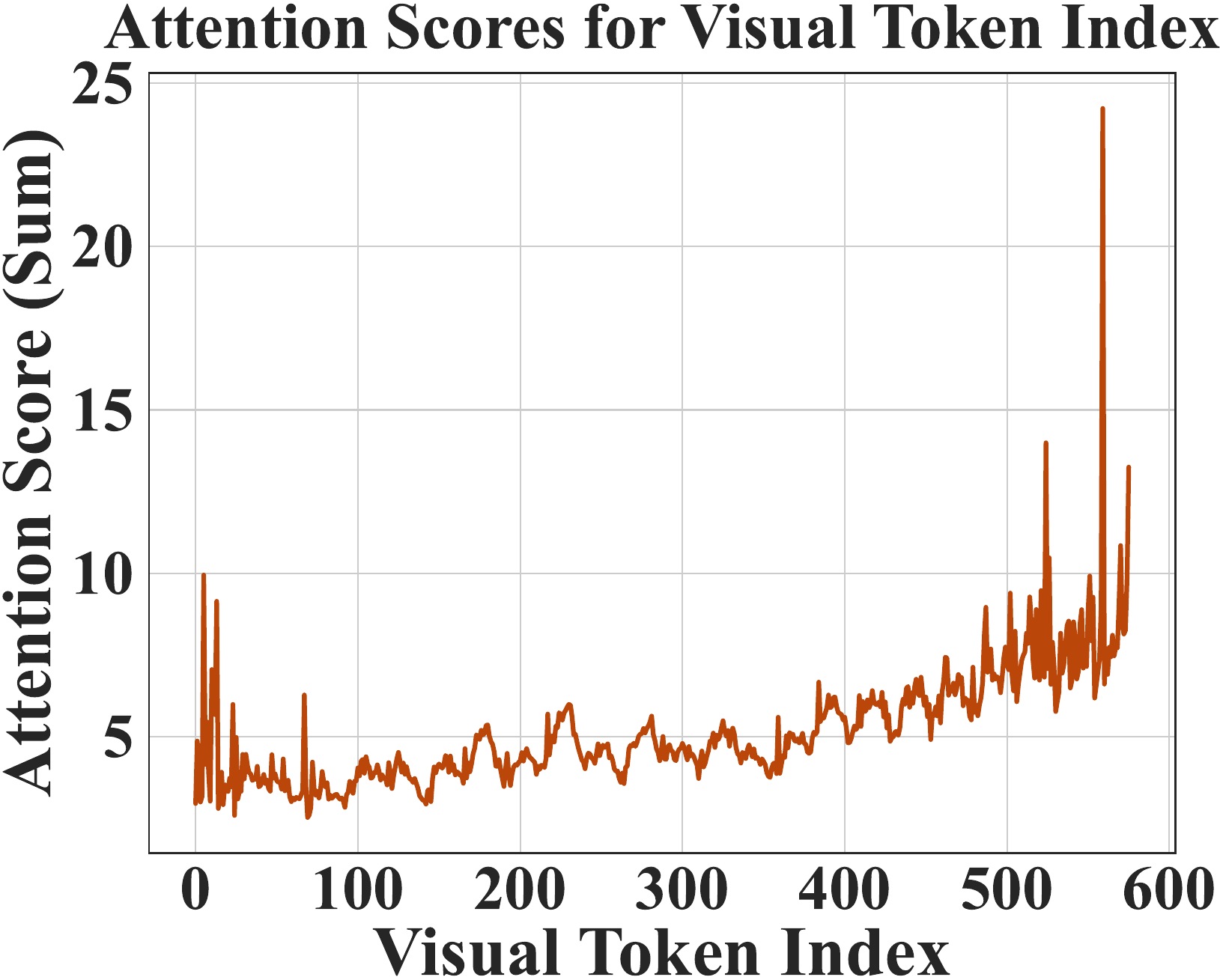}
    }
    \vspace{-4mm}
    \caption{\textbf{Analysis of the distribution of tokens and attention scores over the position of tokens.} Tokens with larger indexes are located at the bottom of images.}
    \vspace{-5mm}
    \label{fig:position_bias}
\end{figure}
\section{Token Pruning Revisited: Are Simple Methods Better?}
When considering token pruning in multimodal large language models, two very basic methods naturally come to mind: random token pruning (hereafter referred to as \textbf{Random}) and token pooling (hereafter referred to as \textbf{Pooling}). Comparison with these two simple baselines is reliable evidence to demonstrate the significance of a well-designed token pruning method, yet has been ignored by most previous works. To address this gap, we investigated these two simple approaches in detail. Specifically, we conducted experiments on multiple widely-used benchmarks under pruning ratios of 75\% and 87.5\%, comparing Random and Pooling\footnote{In the experiments, Pooling specifically refers to applying a pooling operation to the visual tokens at the second layer of the language model. Please refer to the specific implementation in Algorithm~\ref{alg:pooling}.} with several recent token pruning methods (\emph{e.g.}, FastV and SparseVLM). 
As shown in Table~\ref{tab:random_and_pooling}, surprisingly, Random and Pooling outperformed carefully designed methods on nearly $2/3$ benchmarks.
When scaling up to larger models, the experimental results of LLaVA-1.5-13B\footnote{\url{https://huggingface.co/liuhaotian/llava-v1.5-13b}} and Qwen2-VL-72B\footnote{\url{https://huggingface.co/Qwen/Qwen2-VL-72B-Instruct}} in Tables~\ref{tab:random_and_pooling_13B} and~\ref{tab:random_and_pooling_qwen2vl} also demonstrate the superior performance of the simple methods Random and Pooling.
These surprising results shocked us since its inferior performance compared with random selection may demonstrate that we are on the wrong road toward the ideal token pruning methods.

\subsection{Token Distribution: Spatial Uniformity Outperforms Position Bias}
\begin{figure}[!ht]
    \centering
    \includegraphics[width=\linewidth]{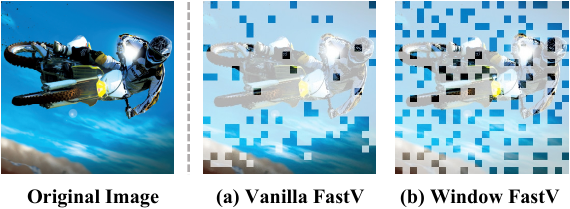}
    \caption{\textbf{Sparse Visualization of Vanilla FastV and Window FastV with 25\% Retained Visual Tokens.}}
    \vspace{-4mm}
    \label{fig:intro_graph}
\end{figure}
\noindent We further explored the underlying reasons behind this phenomenon. Taking FastV~\citep{chen2024image} as an example, this method leverages the attention scores assigned to visual tokens by the last token to evaluate the importance of each visual token, which may introduce the basis for token pruning.

Using 8,910 samples from the POPE dataset, we conducted a statistical analysis of the visual tokens retained by FastV. As illustrated in Figure~\ref{fig:position_bias}, tokens located toward the end of the visual token sequence were assigned significantly higher attention scores and were retained far more frequently than tokens in other positions. This indicates that methods relying on attention scores to select visual tokens inherently suffer from a severe position bias during token reduction. In contrast, tokens retained by Random or Pooling exhibit a naturally uniform spatial distribution. We argue that this spatial uniformity may be the key reason why some existing methods underperform Random and Pooling.

\subsection{Validating the Hypothesis: From Position Bias to Spatial Uniformity}
To validate our hypothesis, we proposed a modification to FastV, introducing a variant called Window FastV. Specifically, we incorporated a sliding window mechanism into the original FastV framework. Within each window, a predetermined reduction ratio and window size were used to select a fixed number of visual tokens. For the specific implementation of Window FastV, please refer to Algorithm~\ref{alg:window_fastv} in Appendix~\ref{app:algorithms}.
Compared to Vanilla FastV, Window FastV ensures the spatial uniformity of the retained tokens, as shown in Figure~\ref{fig:intro_graph}.

We evaluated both Vanilla FastV and Window FastV across eight benchmarks. As shown in Table~\ref{tab:random_and_pooling}, under the setting where 75\% of visual tokens are reduced, Window FastV exhibits an average performance drop that is \textbf{3.4\%} less than that of Vanilla FastV. When adopting a more aggressive reduction ratio (\(\downarrow 88.9\%\)), this gap widens to \textbf{9\%}. These results not only validate our hypothesis but also inspire us to consider strategies that encourage the spatial uniformity of retained tokens when designing token pruning methods. 

To further investigate the impact of token pruning on spatial position understanding, we selected the RefCOCO \citep{yu2016modelingcontextreferringexpressions} dataset, which requires the MLLM to generate a bounding box for a specified object phrase within an image. We consider this dataset to be an effective atomic benchmark for evaluating the spatial understanding capabilities of MLLMs. Our evaluation criterion is that a prediction is considered correct if the Intersection over Union (IoU) between the predicted bounding box and the ground truth area exceeds 0.5. As shown in Table \ref{fig:refcoco_grounding}, compared to conventional tasks, various token pruning methods exhibit a significant degradation in performance when applied to precise object localization. Notably, there is a marked difference between globally uniform attention-based pruning methods (e.g., Window FastV, or even naive approaches like Random and Pooling) and spatially non-uniform strategies (FastV, SparseVLM). This indicates that current token pruning techniques, particularly those that are spatially non-uniform, still possess substantial limitations in comprehending the spatial positioning of objects within images.

\begin{table}[!t]
\centering
\footnotesize
\scalebox{0.68}{
\setlength{\tabcolsep}{2.0pt}
\renewcommand{\arraystretch}{1.25}
\begin{tabular}{c|ccc|ccc|cc|c}
    \toprule[1.5pt]
    \multirow{2}{*}{\textbf{Method}} & \multicolumn{3}{c|}{\textbf{RefCOCO}} & \multicolumn{3}{c|}{\textbf{RefCOCO+}} & \multicolumn{2}{c|}{\textbf{RefCOCOg}} & \multirow{2}{*}{\textbf{Avg.}}  \\
    & \textbf{TestA} & \textbf{TestB} & \textbf{Val} & \textbf{TestA} & \textbf{TestB} & \textbf{Val} & \textbf{Test} & \textbf{Val}  \\
    \hline
    \textcolor{gray}{Vanilla} & \textcolor{gray}{72.3} & \textcolor{gray}{51.5} & \textcolor{gray}{73.3} & \textcolor{gray}{66.3} & \textcolor{gray}{29.7} & \textcolor{gray}{68.3} & \textcolor{gray}{49.5} & \textcolor{gray}{51.5} & \textcolor{gray}{100\%} \\
    \hline
    \textbf{SparseVLMs} & 4.0 & 6.9 & 4.0 & 2.9 & 5.9 & 0.9 & 7.9 & 5.9 & 4.8\% {\color[HTML]{18A6C2} \scriptsize ($\downarrow$ 95.2\%)} \\
    \textbf{Vanilla FastV}   & 20.8 & 13.9 & 27.7 & 17.8 & 8.9 & 26.7 & 19.8 & 14.9 & 18.8\% {\color[HTML]{18A6C2} \scriptsize ($\downarrow$ 81.2\%)} \\
    \textbf{Window FastV} & 22.8 & 22.8 & 25.7 & 18.8 & 7.9 & 18.8 & 20.8 & 23.8 & 20.2\% {\color[HTML]{18A6C2} \scriptsize ($\downarrow$ 79.8\%)} \\
    \textbf{Random} & 22.8 & 27.7 & 32.7 & 17.8 & 13.9 & 32.7 & 20.8 & 16.8 & 23.2\% {\color[HTML]{18A6C2} \scriptsize ($\downarrow$ 76.8\%)} \\
    \textbf{Pooling} & 34.7 & 17.8 & 26.7 & 23.8 & 17.8 & 24.8 & 14.9 & 20.8 & 22.7\% {\color[HTML]{18A6C2} \scriptsize ($\downarrow$ 77.3\%)} \\
    \bottomrule[1.5pt]
\end{tabular}}
\caption{\textbf{Performance Comparison on RefCOCO Series Grounding Tasks}. Evaluation is based on Precision@1, with a reduction ratio of 77.8\%.}
\vspace{-6mm}
\label{fig:refcoco_grounding}
\end{table}

\begin{takeaways}
\ \paragraph{Summary 1.} 
    \emph{The position bias in the distribution of retained visual tokens is a key factor affecting the performance of some existing token pruning methods. This insight suggests that ensuring the spatial uniformity of retained tokens should be an important consideration when designing token pruning strategies.}
\end{takeaways}

\section{Language in  Visual Token Pruning: When and Why Does Language Matter?}

Token pruning methods for multimodal models can be broadly categorized into two types: those guided by textual information (e.g., FastV \cite{chen2024image}, SparseVLM \cite{zhang2024sparsevlm}, MustDrop \cite{liu2024multi}) and those that rely solely on visual information (e.g., FasterVLM \cite{zhang2024clsattentionneedtrainingfree}). While both approaches achieve comparable performance on common benchmarks, however, we hypothesize: Could it be that the importance of language information is not evident simply because there has been a lack of testing on tasks where language information is especially critical? To validate our hypothesis, we select a typical scenario: Visual Haystack.


\subsection{Visual Token Pruning in Strongly Text-Guided Tasks}


Tasks such as Visual Haystack \citep{wu2025visual} (needle-in-a-haystack task on visual scenario) are inherently text-driven. In Visual Haystack task, the MLLM needs to select an image from a set of confusing images with an anchor phrase, and determine whether an object matching a target textual description exists within the selected image. These tasks demand precise alignment between textual and visual modalities. To evaluate the impact of text-guided pruning, we conducted experiments using the LLaVA-1.5-7B model on the VH dataset.

\begin{table}[!h]
\centering
\footnotesize
\scalebox{0.77}{
    \setlength{\tabcolsep}{3.5pt} 
    \renewcommand{\arraystretch}{1.25} 
\begin{tabular}{@{}c|cccccc@{}}
    \toprule[1.5pt]
    \multirow{2}{*}{\textbf{Method}} & \multicolumn{5}{c}{\textbf{\# Input Images} (More images means harder to retrieve)} \\
    & \textbf{Oracle} & \textbf{2} & \textbf{3} & \textbf{5} & \textbf{10} \\ 
    \hline
    \textcolor{gray}{\textbf{LLaVA-1.5-7B}} & \textcolor{gray}{$\text{86.46}_{\pm \text{1.25}}$} & \textcolor{gray}{$\text{70.04}_{\pm \text{1.49}}$} & \textcolor{gray}{$\text{66.18}_{\pm \text{1.58}}$} & \textcolor{gray}{$\text{58.29}_{\pm \text{1.49}}$} & \textcolor{gray}{$\text{53.47}_{\pm \text{1.48}}$} \\ 
    \hline
    \multicolumn{6}{c}{\textbf{Reduction ratio 77.8\%}} \\ 
    \hline
    \textbf{SparseVLM} & $\text{81.26}_{\pm \text{1.11}}$ & $\text{66.14}_{\pm \text{1.54}}$ & $\text{66.54}_{\pm \text{1.33}}$ & $\text{58.22}_{\pm \text{1.51}}$ & $\text{53.99}_{\pm \text{1.65}}$ \\ 
    \textbf{FastV} & $\text{76.30}_{\pm \text{1.36}}$ & $\text{61.17}_{\pm \text{1.56}}$ & $\text{58.34}_{\pm \text{1.61}}$ & $\text{53.39}_{\pm \text{1.51}}$ & $\text{52.06}_{\pm \text{1.63}}$ \\
    \textbf{FastV}$_{\texttt{VIS}}$ & $\text{71.90}_{\pm \text{1.58}}$ & $\text{61.55}_{\pm \text{1.46}}$ & $\text{55.82}_{\pm \text{1.49}}$ & $\text{52.72}_{\pm \text{1.63}}$ & $\text{52.83}_{\pm \text{1.54}}$ \\
    \textbf{Random} & $\text{75.15}_{\pm \text{1.30}}$ & $\text{62.14}_{\pm \text{1.61}}$ & $\text{55.59}_{\pm \text{1.49}}$ & $\text{51.26}_{\pm \text{1.36}}$ & $\text{50.76}_{\pm \text{1.75}}$ \\ 
    \bottomrule[1.5pt]
\end{tabular}}
\caption{\textbf{Performance comparison of different methods on Visual Haystack (VH)}. VH requires MLLMs to select an image from multiple images based on an anchor word and determine the existence of a target word object in the image. FastV$_{\texttt{VIS}}$ means FastV without language information guided.}
\label{tab:visual_haystack}
\end{table}

To validate the importance of text guidance, we modified FastV to operate without textual information and denote it FastV$_{\texttt{VIS}}$. Originally, FastV calculates the importance of visual tokens based on the attention score with the last text token. FastV$_{\texttt{VIS}}$ computes with the last visual token instead, thereby eliminating the influence of text information while preserving the essence of the method. Our results in Table \ref{tab:visual_haystack} show that this modificaiton FastV$_{\texttt{VIS}}$ reveals a significant drop in performance, confirming the importance of leveraging textual cues in strongly text-guided tasks. The comparison of different pruning methods also reveals that approaches utilizing textual information exhibit significantly better overall performance. It is noteworthy that SparseVLM, guided by text information, achieves a compression rate of 77.8\% while maintaining nearly identical accuracy to the uncompressed model, particularly in scenarios with a higher number of confusing images.

However, there are also recent works methods \citep{zhang2024clsattentionneedtrainingfree, liu2025compressionglobalguidancetrainingfree} that perform pruning solely in ViT without textual information and reports better performance than FastV and SparseVLM in common VQA benchmarks. 

Therefore, for tasks with high reliance on language information, pruning strategies should be tailored to incorporate textual guidance effectively, and how to balance the use of linguistic information still requires further research.

\begin{takeaways}
\ \paragraph{Summary 2.} 
    \emph{Text-guided pruning improves performance in text-heavy tasks. Pruning methods should adapt to task needs.}
\end{takeaways}

\section{The \protect\boldmath{$\alpha$} Dilemma: Importance vs. Redundancy in Token Pruning}
\begin{table*}[!h] 
    \vspace{-2mm}
    \centering
    \scalebox{1.0}{
    \setlength{\tabcolsep}{3.5pt} 
    \renewcommand{\arraystretch}{1.0} 
    \footnotesize
    \begin{tabular}{c |c | c  c  c c c c c c c c c}
    \toprule[1.5pt]
    \centering
    \multirow{2}{*}{\textbf{Benchmark}} & \multirow{2}{*}{\textcolor{gray}{\textbf{Vanilla}}} & \multicolumn{11}{c}{\textbf{Balance between Importance and Redundancy $\bm{\alpha}$}}  \\
    \cline{3-13}
    ~ && \cellcolor{mygray}\textbf{0.0} & \cellcolor{mygray}\textbf{0.1} & \cellcolor{mygray}\textbf{0.2} & \cellcolor{mygray}\textbf{0.3} & \cellcolor{mygray}\textbf{0.4} & \cellcolor{mygray}\textbf{0.5} & \cellcolor{mygray}\textbf{0.6} & \cellcolor{mygray}\textbf{0.7} & \cellcolor{mygray}\textbf{0.8} & \cellcolor{mygray}\textbf{0.9} & \cellcolor{mygray}\textbf{1.0} \\
    \hline

    \textbf{MME} & \textcolor{gray}{1862} & 1707 & \colorbox{cyan!10}{1714} & 1711 & 1706 & 1707 & 1711 & 1702 & 1699 & \colorbox{pink!15}{1680} & 1688 & 1689  \\

    \textbf{POPE} & \textcolor{gray}{85.9} & \colorbox{cyan!10}{82.8} & 82.6 & 82.4 & 82.4 & 81.9 & 81.6 & 80.9 & 79.7 & 77.9 & 75.6 & \colorbox{pink!15}{71.8}  \\
    
    \textbf{SQA} & \textcolor{gray}{69.5} & \colorbox{pink!15}{64.8} & 65.2 & 65.2 & 65.1 & 65.1 & 65.3 & 65.3 & 65.2 & 65.5 & \colorbox{cyan!10}{65.7} & 65.3  \\

    \textbf{VQA}$^{\text{Text}}$ & \textcolor{gray}{58.2} & \colorbox{pink!15}{53.6} & 53.8 & \colorbox{cyan!10}{54.8} & 54.0 & 54.1 & 54.3 & 54.3 & 54.5 & 54.4 & 54.2 & 53.6  \\

    \bottomrule[1.5pt]
    \end{tabular}}
    \vspace{-2mm}
    \caption{Performance comparison under different $\bm{\alpha}$ balancing importance and redundancy criteria.}
    \label{tab:similarity_vs_attention}
    \vspace{-4mm}
\end{table*}
In this section, we systematically analyze the fundamental tension in token pruning for multimodal large language models: \emph{should we prioritize removing redundant tokens to preserve structural patterns, or eliminate less important tokens to maintain predictive capacity?}

\subsection{Redundancy Criteria} 
\label{sec:Redundancy_criteria}
This criterion adopts a \emph{task-agnostic} perspective, focusing exclusively on input patterns. The core objective is to eliminate redundant tokens while preserving the input's \emph{structural integrity} and minimizing information loss - analogous to finding the minimal sufficient statistics in information theory.

Through the lens of mutual information~\citep{latham2009mutual}, we formulate this as maximizing information preservation between original tokens $\mathbf{X}$ and retained tokens $\mathbf{X'}$: 
\begin{equation}
    \max_{\mathcal{P}} I(\mathbf{X}; \mathbf{X'}) = \mathcal{H}(\mathbf{X}) - \mathcal{H}(\mathbf{X}|\mathbf{X'}),
\end{equation}
where $\mathcal{P}$ denotes the pruning operator. This ensures the $\mathbf{X'}$ retains maximal dependence on $\mathbf{X}$ under length constraint $\|\mathbf{X'}\| = \|\mathbf{X}\| - \Delta L$. The formulation directly connects to the \emph{compression phase} of the information bottleneck principle~\citep{tishby2000information}, where $\mathcal{P}^*$ solves:
\begin{equation}
    \mathcal{P}^* = \argmin_{\mathcal{P}} \|\mathbf{X'}\| \quad \text{s.t.} \ I(\mathbf{X};\mathbf{X'}) \geq \gamma,
\end{equation}
with $\gamma$ as the minimal acceptable mutual information. This preserves structural patterns without task-specific considerations.

\subsection{Importance Criteria} 
\label{sec:Important_criteria}
In contrast, this \emph{task-oriented} criterion explicitly considers the target output $\mathbf{Y}$. The goal shifts to preserving tokens critical for \emph{prediction accuracy}, formalized through predictive sufficiency: 
\begin{equation}
    I(\mathbf{X'}; \mathbf{Y}) \geq I(\mathbf{X}; \mathbf{Y}) - \epsilon,
\end{equation}
where $\epsilon$ is the tolerable information loss. Expanding via the chain rule:
\begin{equation}
    \underbrace{I(\mathbf{X}; \mathbf{Y})}_{\text{Original}} = \underbrace{I(\mathbf{X'}; \mathbf{Y})}_{\text{Pruned}} + \underbrace{I(\mathbf{X}\setminus \mathbf{X'}; \mathbf{Y}|\mathbf{X'})}_{\text{Discarded}}.
\end{equation}
The bound $I(\mathbf{X}\setminus \mathbf{X'}; \mathbf{Y}|\mathbf{X'}) \leq \epsilon$ implies that discarded tokens provide negligible additional information about $\mathbf{Y}$ when conditioned on retained tokens. This captures the essence of importance - truly critical tokens contain non-decomposable predictive information. 

The task dependence manifests in the information plane:
\begin{equation}
    \mathcal{R}(\beta) = \max_{\mathbf{X'}} \left[ I(\mathbf{X'};\mathbf{Y}) - \beta^{-1}I(\mathbf{X};\mathbf{X'}) \right],
    \label{eq:banlance}
\end{equation}
where $\beta$ controls redundancy-importance tradeoff. 

\subsection{Empirical Validation of Adaptive Criteria Balancing}
\label{sec:Empirical_Validation}
Building on Eq.~\ref{eq:banlance}, we implement an adaptive scoring mechanism with tunable parameter $\alpha$: 
\begin{align}
    \text{Score}(x_i) = \nonumber \\
    &\hspace{-3em} \alpha \cdot \hspace{-0.7em}\underbrace{I(x_i; \mathbf{Y}|x_{\setminus i})}_{\text{Predictive Criticality}} \hspace{-0.7em}+ (1-\alpha) \cdot \underbrace{[1 - I(x_i; \mathbf{X}_{\setminus i})]}_{\text{Pattern Uniqueness}}.
    \label{eq:alpha_balance}
\end{align}
Here $I(x_i; \mathbf{Y}|x_{\setminus i})$ measures a token's unique predictive value, while $1 - I(x_i; \mathbf{X}_{\setminus i})$ quantifies its pattern distinctiveness.

Specifically, FastV is a typical token pruning method that follows the importance criterion by selecting important visual tokens based on the attention scores of the last token in the sequence. We modify this approach by introducing a redundancy criterion, which calculates the cosine similarity between each visual token and the last token to derive a similarity score\footnote{Notably, since the similarity score and attention score are on different scales, we apply min-max normalization to both before computing the final score.}. Ultimately, the final score in Eq.~\ref{eq:alpha_balance} is obtained by balancing these two metrics with a parameter $\alpha$.
Our experiments results in Table~\ref{tab:similarity_vs_attention} reveal two key insights: 

\begin{itemize}[leftmargin=10pt, topsep=0pt, itemsep=1pt, partopsep=1pt, parsep=1pt]
    \item \textbf{Perception-Dominant Tasks} (MME, POPE) achieve peak performance at $\alpha=0.1$ and $0.0$, respectively, favoring redundancy-first pruning to maintain structural integrity ($\uparrow I(\mathbf{X};\mathbf{X'})$).
    
    \item \textbf{Knowledge-Intensive Tasks} (SQA, VQA$^\text{Text}$) achieve optimal performance with $\alpha=0.8\sim0.9$, favoring importance-first pruning to enhance semantic coherence ($\uparrow I(\mathbf{X'};\mathbf{Y})$).
\end{itemize}
\begin{takeaways}
\ \paragraph{Summary 3.}
\emph{Prune by task: Redundancy-first preserves structural fidelity for perception tasks, while importance-first prioritizes predictive power for knowledge reasoning.}
\end{takeaways}



\section{Limitations and Challenges in Token Pruning Evaluation}
Token pruning has emerged as a promising technique to improve the efficiency of MLLMs. However, despite its potential, the evaluation of token pruning methods remains fraught with challenges. 
In this section, we critically examine two key issues that hinder the accurate and meaningful assessment of token pruning techniques: \textbf{(i)} the over-reliance on FLOPs as a proxy for speed gains, and \textbf{(ii)} the failure to account for training-aware compression in some advanced MLLMs. 
We argue that addressing these challenges is crucial for developing more robust and reliable token pruning approaches.

\subsection{Beyond FLOPs: Shifting the Focus to Actual Latency Gains}
\begin{table}[!h]
    \centering
    \setlength{\tabcolsep}{2.0pt}
    \renewcommand{\arraystretch}{1.2}
    \footnotesize
    \scalebox{0.7}{
    \begin{tabular}{@{}lcccccccc}
        \toprule[1.2pt]
         \multirow{2}{*}{\textbf{Methods}} & \multirow{2}{*}{\textbf{Tokens $\downarrow$}} & \textbf{Latency $\downarrow$} & \multirow{2}{*}{\textbf{FLOPs $\downarrow$}} & \textbf{KV Cache $\downarrow$}  & \multirow{1}{*}{\textbf{POPE $\uparrow$}}   \\
           && \textbf{(Min:Sec)} & & \textbf{(MB)} & \textbf{(F1-Score)} \\
         \midrule
         \textcolor{gray}{Vanilla LLaVA-Next-7B} & \textcolor{gray}{2880} & \textcolor{gray}{36:16} & \textcolor{gray}{100\%} & \textcolor{gray}{1512.1} &\textcolor{gray}{86.5}  \\
         \hspace{0.5em} + FastV & 320 & 18:17 & \textbf{12.8\%} & 168.0 & 78.3  \\
         \hspace{0.5em} + SparseVLM & 320 & 23:11 & 15.6\% & 168.0 & 82.3  \\
          \hspace{0.5em} + MustDrop & 320 & 23:40  & 11.5\% & 168.0 & 82.1 \\
       
        \bottomrule[1.2pt]
    \end{tabular}}
    \vspace{-2mm}
    \caption{Inference costs of the number of tokens, Total-Time, FLOPs, and KV Cache Memory.}
    \label{tab:efficiency}
    \vspace{-6mm}
\end{table}
\noindent\textbf{Phenomenon.} Many existing token pruning approaches tend to measure the speedup of their methods by calculating or estimating the reduction in FLOPs resulting from token reduction, or even directly using the token reduction ratio as a metric. However, can FLOPs or token reduction ratios truly reflect the actual acceleration achieved?

To investigate this question, we examined the speedup effects reported by several works. Our findings reveal that even when different methods exhibit identical or similar reduction ratios and FLOPs, their measured speeds can vary significantly. 
Table~\ref{tab:efficiency} presents the efficiency-related experimental results of these methods on LLaVA-Next-7B\footnote{\url{https://huggingface.co/liuhaotian/llava-v1.6-vicuna-7b}}. Specifically, under the same setting, SparseVLM's FLOPs are only \textbf{2.8\%} higher than those of FastV, yet its latency is \textbf{26.8\%} greater. 
This strongly suggests that relying on FLOPs to evaluate acceleration effects of proposed methods is inadequate. When assessing speed gains, it is imperative to \emph{shift our focus to actual latency measurements}.

\noindent\textbf{Reason.} We also conducted a detailed analysis of the design intricacies of the three methods to uncover the underlying reasons for their performance differences. Specifically, FastV, SparseVLM, and MustDrop all fail to support the efficient Flash Attention operator~\citep{dao2022flashattention, dao2023flashattention2}, as they rely on the complete attention map to select visual tokens. However, FastV performs token pruning in only one layer of the language model, whereas the other two methods conduct pruning across four layers. This implies that, compared to FastV, these methods have more layers that are forced to use the traditional attention operator with $O(N^2)$ memory costs. This could be one of the key factors contributing to their slower speeds.
Additionally, performing pruning layer by layer requires more complex operations to handle token selection. If the runtime overhead introduced during this stage becomes significant, it may offset the speed gains achieved by shortening the token sequence. Moreover, some of the transformer layers where these methods perform pruning are located deeper within the model. Pruning tokens in such deep layers may yield limited benefits, as the impact of token reduction diminishes at later stages of the network.

\noindent\textbf{Appeal.} This insight motivates us to consider the compatibility with efficient attention operators when designing token pruning methods. Additionally, it encourages us to implement the token pruning process as early as possible in the shallow layers using simpler approaches, avoiding the risk of excessive runtime overhead that could otherwise overshadow the intended acceleration benefits.
\begin{takeaways}
\ \paragraph{Summary 4.} 
    \emph{(i) FLOPs are not a reliable metric for evaluating speed gains; greater emphasis should be placed on actual latency. (ii) We advocate for the implementation of token pruning in the shallow layers of MLLMs using simple or efficient operations, while ensuring compatibility with Flash Attention.}
\end{takeaways}

\subsection{The Overlooked Role of Training-Aware Compression in MLLMs}
\begin{table}[!h]
    \vspace{-3mm}
    \centering
    \resizebox{0.48\textwidth}{!}{\setlength{\tabcolsep}{1.7pt}
    \renewcommand{\arraystretch}{1.1}
    {\begin{tabular}{p{2.35cm} | c c c c c c c | >{\centering\arraybackslash}p{2.0cm}}
        \toprule[1.5pt]
        \textbf{Method} & \textbf{GQA} & \textbf{MMB} & \textbf{MMB-CN} & \textbf{MME} & \textbf{POPE} & \textbf{SQA} & \textbf{VQA}$^{\text{Text}}$ & \makecell[c]{\textbf{Avg}.}\\
        \hline
        \rowcolor{mygray}
        Qwen2-VL-7B & \multicolumn{8}{c}{\textit{Upper Bound, All Tokens} \ $\textbf{(100\%)}$}   \\
        \textcolor{gray}{Vanilla} & \textcolor{gray}{62.2} & \textcolor{gray}{80.5} & \textcolor{gray}{81.2} & \textcolor{gray}{2317} & \textcolor{gray}{86.1} & \textcolor{gray}{84.7} & \textcolor{gray}{82.1} & \multirow{1}*{\textcolor{gray}{100\%}} \\
        \hline

        \rowcolor{mygray}
        Qwen2-VL-7B & \multicolumn{8}{c}{\textit{Token Reduction} \ $\fg{(\downarrow 66.7\%)}$}   \\
        \hspace{0.2em} + FastV & 58.0 & 76.1 & 75.5 & 2130 & 82.1 & 80.0 & 77.3 & 94.0\% {\color[HTML]{18A6C2} \scriptsize (-6.0\%)} \\
        \hspace{0.2em} + \multirow{1}*{FastV$^\dagger$} & 61.9 & 80.9 & 81.3 & 2296 & 86.2 & 84.6 & 81.7 & 99.8\% {\color[HTML]{18A6C2} \scriptsize (-0.2\%)}  \\
        \hline

        \rowcolor{mygray}
        Qwen2-VL-7B & \multicolumn{8}{c}{\textit{Token Reduction} \ $\fg{(\downarrow 77.8\%)}$}\\
        \hspace{0.2em} + \multirow{1}*{FastV} & 56.7 & 74.1 & 73.9 & 2031 & 79.2 & 78.3 & 72.0 & 91.0\% {\color[HTML]{18A6C2} \scriptsize (-8.0\%)} \\
        \hspace{0.2em} + \multirow{1}*{FastV$^\dagger$} & 61.9 & 80.8 & 81.2 & 2300 & 86.1 & 86.4 & 81.4 & 100.0\% {\color[HTML]{18A6C2} \scriptsize (0.0\%)}  \\

        \hline

        \rowcolor{mygray}
        Qwen2-VL-7B & \multicolumn{8}{c}{\textit{Token Reduction} \ $\fg{(\downarrow 88.9\%)}$}\\
        \hspace{0.2em} + \multirow{1}*{FastV} & 51.9 & 70.1 & 65.2 & 1962 & 76.1 & 75.8 & 60.3 & 84.0\% {\color[HTML]{18A6C2} \scriptsize (-16.0\%)}  \\
        \hspace{0.2em} + \multirow{1}*{FastV$^\dagger$} & 61.9 & 81.1 & 81.0 & 2289 & 86.2 & 84.4 & 81.3 & 99.6\% {\color[HTML]{18A6C2} \scriptsize (-0.4\%)}  \\

        \bottomrule[1.5pt]
	\end{tabular}}}
    \vspace{-3mm}
	\caption{Comparative Experiments on Qwen2-VL-7B.}
    \vspace{-3mm}
    \label{app_tab:qwen2vl}
 
\end{table}
In recent years, some of the latest MLLMs have adopted various advanced techniques during the training phase to enhance their efficiency. For instance, Qwen2-VL employs token merging strategy during training, consolidating four adjacent patches into a single visual token. Similarly, MiniCPM-V-2.6 incorporates a learnable query within its resetting module, mapping variable-length segment features into more compact representations.

This raises an intriguing question: \emph{If MLLMs already implement training-aware compression techniques, should we take this into account when designing and evaluating token pruning methods for the inference stage?} Given that the visual tokens encoded by these models possess higher information density, removing the same number of visual tokens could result in greater information loss compared to traditional approaches.

To this end, we selected a representative MLLM that employs training-aware compression, Qwen2-VL-7B-Instruct\footnote{\url{https://huggingface.co/Qwen/Qwen2-VL-7B-Instruct}} and conducted a series of experimental analyses. Specifically, we applied FastV in two sets of experiments: one disregarding the token compression performed during Qwen2-VL's training phase, and the other taking it into account:
Let $\mathcal{P}$ denote the original number of image patches, and after processing with PatchMerger, the number of visual tokens $\mathcal{V}$ is:
\begin{equation}
    \mathcal{V} = \frac{\mathcal{P}}{\text{TACR}},
\end{equation}
where TACR means training-aware compression ratio and the value is 4. 

Finally, the Token Reduction Rate (TRR) can be formally defined as:
\begin{equation}
    \text{TRR}(\text{FastV}^\dagger) \triangleq 
    \underbrace{\text{TACR}}_{\text{Training-aware}} \times 
    \underbrace{\text{TFRR}}_{\text{Training-free}}.
    \vspace{-1mm}
\end{equation}

Surprisingly, as shown in Table~\ref{app_tab:qwen2vl}, we find that when taking training-aware compression into account, the same token pruning method achieves performance on par with the vanilla model across multiple benchmarks, even under varying reduction ratios.
This observation prompts us to reflect: \emph{perhaps more research effort should be directed toward training-aware token compression techniques}. Even in cross-model comparisons, such as between LLaVA-1.5-7B (Vanilla FastV) in Table~\ref{tab:random_and_pooling}, which does not employ training-aware compression, and Qwen2-VL-7B-Instruct (FastV$^\dagger$), the latter clearly demonstrates less performance degradation. 
\begin{takeaways}
\ \paragraph{Summary 5.} 
    \emph{Training-aware token compression techniques deserve more research attention due to their potential for delivering superior performance guarantees.}
\end{takeaways}

\section{Conclusion}


Our systematic investigation into token pruning for MLLMs reveals several critical yet overlooked issues. While existing methods prioritize attention-based scoring and language-guided strategies, we demonstrate that naive spatial uniformity, achieved through random selection or pooling, often outperforms complex designs due to inherent positional biases in visual tokens. Notably, the effectiveness of linguistic guidance depends on task alignment: it enhances performance in text-driven scenarios through cross-modal attention but risks degradation in vision-centric tasks. 
From an information-theoretic perspective, we shed light on the core principles of token pruning, \emph{i.e.,} the pursuit of structural integrity versus prediction accuracy. Furthermore, we challenge the conventional reliance on FLOPs for efficiency evaluation, showing that latency serves as a more practical and meaningful metric. 
These findings provide a refined framework to guide the development of future token pruning methods, balancing simplicity, effectiveness, and task-specific adaptability.



\section{Limitations}
Our experiments and analyses have been primarily conducted on LLaVA, LLaVA-Next, and Qwen2-VL. While these multimodal large language models are highly representative, our exploration should be extended to a broader range of model architectures. Such an expansion would enable us to uncover more intriguing findings and gain more robust and comprehensive insights. Additionally, we should apply our analytical framework and experimental evaluations to models of varying sizes, ensuring that our conclusions are not only diverse but also applicable across different scales of architecture.
\clearpage
\bibliography{custom}

\clearpage
\appendix

\section{Future Works}
In this work, we have conducted an in-depth exploration of a series of issues related to token pruning. These include the position bias problem inherent in methods based on attention scores, the guiding role of linguistic information during token pruning, the importance and redundancy of tokens, as well as certain limitations in the evaluation of token pruning methods.
Looking ahead, we plan to further expand the scope of our research by considering whether token pruning or token merging should be prioritized in the context of token reduction. Additionally, we aim to evaluate and analyze various token reduction methods on more challenging OCR benchmarks~\citep{zhang2024ocr,ouyang2024omnidocbenchbenchmarkingdiversepdf}, particularly datasets featuring rich-text OCR images.
This future work will not only deepen our understanding of token reduction strategies but also provide valuable insights into their practical applications in complex scenarios.

\section{Algorithms}\label{app:algorithms}
In this section, we present some core algorithms for the methods mentioned in the main text. Vanilla FastV (Algorithm~\ref{alg:fastv_core}) selects tokens with the highest attention scores for retention. Reverse FastV (Algorithm~\ref{alg:reverse_fastv}) modifies this strategy by selecting tokens with the lowest attention scores instead. Window FastV (Algorithm~\ref{alg:window_fastv}) introduces a spatially-aware token selection mechanism by dividing the image tokens into local windows and performing token selection within each window. Finally, Pooling (Algorithm~\ref{alg:pooling}) applies a pooling operation over token grids to retain a structured subset of tokens, ensuring spatial consistency.

\begin{algorithm}[!h]
\caption{Vanilla FastV}
\label{alg:fastv_core}
\begin{algorithmic}[1]
\Require Input token sequence $X \in \mathbb{R}^{L \times d}$, image token range $[s,e]$, retention ratio $r$
\Ensure Compressed sequence representation $X'$

\State Initialize layer parameters $\{W_i\}_{i=1}^N$
\For{layer $l = 1$ \textbf{to} $N$}
    \If{$l = K-1$}
        \State Compute attention matrix $A$
        \State Record global attention scores $\alpha = \text{mean}(A)[s:e]$
    \ElsIf{$l = K$}
        \State Select top-k indices $I = \text{topk}(\alpha, \lfloor (e-s)r \rfloor)$
        \State Construct retention indices $\mathcal{I} = [0:s) \cup I \cup [e:L]$
        \State Compress sequence $X' = X[\mathcal{I},:]$
        \State Update attention mask $M' = M[\mathcal{I},\mathcal{I}]$
    \Else
        \State Regular Transformer computation
    \EndIf
\EndFor
\end{algorithmic}
\end{algorithm}

\begin{algorithm}[t]
\caption{Reverse FastV}
\label{alg:reverse_fastv}
\begin{algorithmic}[1]
\Require Input token sequence $X \in \mathbb{R}^{L \times d}$, image token range $[s,e]$, retention ratio $r$
\Ensure Compressed sequence representation $X'$

\State Initialize layer parameters $\{W_i\}_{i=1}^N$
\For{layer $l = 1$ \textbf{to} $N$}
    \If{$l = K-1$}
        \State Compute attention matrix $A$
        \State Record global attention scores $\alpha = \text{mean}(A)[s:e]$
        \State $\alpha = -\alpha$ \Comment{\textcolor{blue}{Difference from Vanilla FastV}}
        
    \ElsIf{$l = K$}
        \State Select top-k indices $I=\text{topk}(-\alpha, \lfloor (e-s)r \rfloor)$ 
        \State Construct retention indices $\mathcal{I} = [0:s) \cup I \cup [e:L]$
        \State Compress sequence $X' = X[\mathcal{I},:]$
        \State Update attention mask $M' = M[\mathcal{I},\mathcal{I}]$
    \Else
        \State Regular Transformer computation
    \EndIf
\EndFor
\end{algorithmic}
\end{algorithm}

\begin{algorithm}[t]
\caption{Window FastV}
\label{alg:window_fastv}
\begin{algorithmic}[1]
\Require Input sequence $X \in \mathbb{R}^{L \times d}$, image region $\Omega=[s,e]$, window size $(h,w)$ 
\Ensure Compressed sequence representation $X'$

\State Initialize layer parameters $\{W_i\}_{i=1}^N$
\For{layer $l = 1$ \textbf{to} $N$}
    \If{$l = K-1$}
        \State Compute attention matrix $A$
        \State Record global attention scores $\alpha = \text{mean}(A)[s:e]$
    \ElsIf{$l = K$}
        \State Reshape the image region into a 2D grid $\Gamma \in \mathbb{R}^{h\times w}$
        \State Divide the grid into window patches $\{\mathcal{W}_{ij}\}_{i=1,j=1}^{m,n}$, where $\mathcal{W}_{ij} \subset \Gamma$
        
        \For{each window $\mathcal{W}_{ij}$}
            \State Compute local attention scores $A_{ij} = \text{mean}(\alpha[\mathcal{W}_{ij}])$
            \State Select local top-k indices $I_{ij} = \text{topk}(A_{ij}, mn)$
            \State Convert local indices to global coordinates $\mathcal{G}_{ij} = \text{loc2glob}(I_{ij})$
        \EndFor
        \State Aggregate all window indices $\mathcal{I} = \bigcup_{i,j} \mathcal{G}_{ij}$
        \State Construct the retained sequence:
        $
        X' = X\left[[0:s) \cup \mathcal{I} \cup [e:L], :\right]
        $
    \Else
        \State Regular Transformer computation
    \EndIf
\EndFor
\end{algorithmic}
\end{algorithm}

\begin{algorithm}[t]
\caption{Pooling}
\label{alg:pooling}
\begin{algorithmic}[1]
\Require Input sequence $X \in \mathbb{R}^{L \times d}$, image region $\Omega=[s,e]$, window size $a \times a$
\Ensure Compressed sequence representation $X'$
\For{layer $l = 1$ \textbf{to} $N$}
    \If{$l = K$}
        \State Extract image tokens: $X_{img} = X[s:e,:]$
        \State Reshape into a 2D grid: $F \in \mathbb{R}^{h \times w \times d}$ \\
        \Comment{Where $h \times w = e-s$}
        \State Perform window pooling:
        $$
        \hat{F} = \text{Pool}(F, a, \rho) \in \mathbb{R}^{(h/a) \times (w/a) \times d}
        $$
        \State Construct index mapping:
        $$
        \mathcal{M} = \{(i,j) \mapsto \mathop{\text{argmax}}\limits_{(p,q) \in \mathcal{W}_{ij}} \|F[p,q,:]\|_1\}
        $$
        \State Build the retained index set:
        $$ \mathcal{I} = \{s + \mathcal{M}(k) | \forall k \in [1, (h/a)(w/a)]\} $$
        \State Generate the compressed sequence:
        $$
        X' = X\left[[0:s) \cup \mathcal{I} \cup [e:L], :\right]
        $$
    \Else
        \State Regular Transformer computation
    \EndIf
\EndFor
\end{algorithmic}
\end{algorithm}

\section{Dataset}\label{app:dataset details}

In this section, we introduce the content of the datasets used, as well as the input and output formats in Table \ref{tab:datasets}.

\begin{table*}[t]
    \centering
    \footnotesize
    \renewcommand{\arraystretch}{1}
    \setlength\tabcolsep{6pt}
    \begin{tabular}{lllll}
        \toprule
        \bf{Type}                                  & \bf{Dataset}                         & \bf{Brief Description}                                     & \bf{Input}                                                        & \bf{Output}                                            \\
        \midrule
        \multirow{10}{1.8cm}{Visual Understanding} & \multirow{1}{*}{VQA$^\text{Text}$}            & \multirow{1}{*}{Rich textual viusal QA}                    & \multirow{1}{*}{Single image and question}          & Question Answer                                        \\
        \cmidrule{2-5}
                                                   & \multirow{1}{*}{VQA V2}              & \multirow{1}{*}{Open-ended viusal perception}              & \multirow{1}{*}{Single image and question}     & Question Answer                                        \\
        \cmidrule{2-5}
                                                   & \multirow{1}{*}{ScienceQA}           & \multirow{1}{*}{Natural and social science QA}             & \multirow{1}{*}{Single image and question}     & Question Answer                                        \\
        \cmidrule{2-5}
                                                   & \multirow{1}{*}{POPE}                & \multirow{1}{*}{Object hallucination evaluation}           & \multirow{1}{*}{Single image and question}     & Question Answer                                        \\
        \cmidrule{2-5}
                                                   & \multirow{1}{*}{GQA}                 & \multirow{1}{*}{Visual scene understanding}                & \multirow{1}{*}{Single image and question}     & Question Answer                                        \\
        \cmidrule{2-5}
                                                   & \multirow{1}{*}{MMBench}             & \multirow{1}{*}{Perception and reasoning tasks}            & \multirow{1}{*}{Single image and question}     & Question Answer                                        \\
        \cmidrule{2-5}
                                                   & \multirow{1}{*}{MME}                 & \multirow{1}{*}{Perceptual ability evaluation}             & \multirow{1}{*}{Single image and question}     & Question Answer                                        \\
        \midrule
        \multirow{2}{1.8cm}{Object Recognition}    & \multirow{2}{1.5cm}{Visual Haystack} & \multirow{2}{*}{Visual need-in-a-haystack}                 & \multirow{2}{3.5cm}{Multiple images, an image phrase and an object phrase}                                                                  & \multirow{2}{2.3cm}{Existence of specified objects}    \\
                                                   &                                      &                                                            &                     &                                                        \\
        \midrule
        \multirow{2}{1.8cm}{Grounding}             & \multirow{2}{*}{RefCOCO}             & \multirow{2}{*}{Phrase object localizing}                  & \multirow{2}{3.5cm}{1 image and referring phrase of an object} & \multirow{2}{2.5cm}{Bounding box of the specified object} \\
                                                   &                                      &                                                            &                                                                   &                                                        \\
        \bottomrule
    \end{tabular}
    \caption{The datasets we use for benchmarking.}
    \label{tab:datasets}
\end{table*}

\end{document}